# The NAO Backpack: An Open-hardware Add-on for Fast Software Development with the NAO Robot

Matías Mattamala*, Gonzalo Olave*, Clayder González, Nicolás Hasbún and Javier Ruiz-del-Solar

Department of Electrical Engineering, Universidad de Chile
Advanced Mining Technology Center (AMTC)
Av. Tupper 2007, Santiago, Chile

`{mmattamala,golave,jruizd}@ing.uchile.cl`

**Abstract.** We present an open-source accessory for the NAO robot, which enables to test computationally demanding algorithms in an external platform while preserving robot's autonomy and mobility. The platform has the form of a backpack, which can be 3D printed and replicated, and holds an ODROID XU4 board to process algorithms externally with ROS compatibility. We provide also a software bridge between the B-Human's framework and ROS to have access to the robot's sensors close to real-time. We tested the platform in several robotics applications such as data logging, visual SLAM, and robot vision with deep learning techniques. The CAD model, hardware specifications and software are available online for the benefit of the community.

## 1   Why would the NAO need a backpack?

Let us start with a short story: *some weeks ago, Arya, one of the NAO robots in our lab, became tired of always having the same behaviors and applying the same logic for all her problems, recognizing lines, blobs and other robots using the same old-fashioned algorithms. After some hours reflecting on it, she horrified realized that it was impossible for her to figure out other computer vision paradigms; moreover, she had become aware of her own existence and physical limitations. There was knowledge beyond its restricted understanding, solutions unreachable for its limited capabilities.* This story, despite fictional, hides a real issue in the Standard Platform League (SPL): Developments are limited by the constraints of the standard platform itself [1]. Softbank's NAO robot currently used in the SPL presents several advantages over other humanoid platforms -for instance, the number of degrees of freedom or its price-, but possess strong limitations in terms of processing power, having only a single physical core (2 virtual). This issue demands the use of custom frame-works heavily optimized for the platform in order to be used in *real-time*[1] applications such as robot-soccer. Hence, huge advances and software solutions from the robotics community, mainly developed within the widely-used ROS framework, are not easily transferable or applicable in the robot soccer domain. Analogously, it is difficult for the SPL community to provide solutions for the other robotics communities apart of efficient algorithms to solve domain-specific problems.

---

\* These authors contributed equally to this work

In this work, we face the previous issue by presenting a new open-hardware accessory for the NAO which allows us to test algorithms in a powerful, ROS-compatible platform, as well as to anticipate future developments for the league: The **NAO Backpack**. This device lets the NAO robot carry more powerful hardware -an ODROID XU4- while preserving autonomy and mobility, avoiding the issues of using external hardware via Wi-Fi (high latency) or long Ethernet cables that limits the robot's movements, allowing to test algorithms in a realistic setup.

Here we provide the guidelines to reproduce the NAO backpack, its physical design and software involved. In Section 2 we first show a brief review of commercial and open-source accessories for the NAO. In Section 3 we describe the hardware and software components of the NAO backpack. In Section 4 we show some applications of the backpack for real-time dataset recording, visual SLAM and deep learning. Finally, in Section 5, we finish by covering some future backpack improvements and challenges for the community.

## 2 A Brief Review of NAO Accessories

To the best of our knowledge, the range of accessories for the NAO robot is very limited. There are a few commercial products available, but mainly focused on extending its *exterioceptive* capabilities or facilitating its use in human environments. Some of them are listed as follows:
- Aldebaran Robotics' Laser Head [3]: An official but discontinued Hokuyo laser rangefinder for mapping applications.
- Robots Lab's NAO Car [4]: A NAO-sized BMW Z4 electric car with laser mapping capabilities
- Robots Lab's Docking station [5]: A fancy-looking seating to charge the NAO.

We also revised some of the most popular 3D printing projects websites (GrabCAD, Thingiverse and Autodesk 123D) looking for NAO projects, and we found the following:
- NAO Helmet for Microsoft Kinect [6]: A 3D printed helmet for a Kinect model.
- NAO Helmet for Asus Xtion [7]: 3D printed helmet for the Asus RGB-D sensor.
- NAO Bag for ODROID U3 [8]: This project designed by K. Chatzilygeroudis is comparable to ours, so we discuss the differences below.

The Chatzilygeroudis' NAO Bag was designed to be attached on top of the current battery cover; however, there are no specifications about power supply, connectivity, or benchmarking. This bag was after used by Canzobre *et al*. [9], who improved the discussion by providing references about similar works, as well as providing comparisons between different CPUs (ODROID XU3 and U3), and commercially available power supplies. They also propose different mounting configurations to hold both the CPU and the batteries, and showed an application of robot mapping with RGB-D sensors.

---

[1] In the robot soccer domain, we understand as *real-time* tasks that can be processed in a frequency close to camera rate, i.e. 15-30 Hz.

This work has several similitudes to ours, but the differences are significant. On the contrary to Canzobre et al. whose mounting designs are not openly available, we provide a full open backpack design, to be fabricated easily with current rapid prototyping tools; our design can carry both the external processing units as well as the battery, and works as a replacement for the battery cover. Despite not providing comparisons with different processing units and power supplies, we propose a set of tested components and provide the physical parameters of the backpack for that configuration, such as the mass, center of mass and inertia. Furthermore, we also provide software bridges to communicate a NAO running the B-Human (BH) framework, widely used in the SPL [2], with the backpack. The bridge allows us to exploit the efficiency of the BH software while taking advantage of the ROS community developments.

In summary, few accessories are available for the NAO nowadays. The expensive price for home users as well as the no hardware modification rule of the SPL [1] can explain part of this situation. Some previous attempts to extend the processing capabilities of the NAO have been tried before, but most of them have prevailed closed within research communities. With this work, we hope to contribute to spread these ideas as well as to motivate discussions about the current limitations in the SPL. The NAO Backpack

### 2.1 Mechanical Design

We wanted the backpack to be easily attached and detached from the robot, so we base the design on the plastic rear cover of the battery maintaining the through holes for the screws to fit in and hold the backpack. Since we are concerned about autonomy, a hard requirement was to fit the ODROID XU4, a 2-cell LiPo battery, and a voltage regulator. For more details refer the next section.

**Center of Mass and Inertia.** For our purposes, it is mandatory to give inertial information about the backpack's physical parameters to generate the appropriate matrices for calculating the dynamic movements of the Nao robot, in both simulated and real tests. The units of the *inertial matrix* $I_0$, *mass* $M_0$ and *center of mass* $C_0$ are all in MKS system, calculated with respect to the coordinate system showed in the Figure 1b. All these parameters were computed by considering the real distribution of components in the backpack.

$$I_0 = \begin{pmatrix} 5.66e^{-4} & 3.74e^{-6} & -2.13e^{-4} \\ 3.74e^{-6} & 6.46e^{-4} & -9.76e^{-6} \\ -2.13e^{-4} & -9.76e^{-6} & 8.17e^{-5} \end{pmatrix} \quad M_0 = 0.2074 \quad C_0 = \begin{pmatrix} -0.0197 \\ 0 \\ 0.052 \end{pmatrix}$$

**Construction Aspects.** Because one objective of the backpack is to be open hardware, we use 3D printing for its construction. The printer we used is the XYZ DaVin-ci Jr 1.0 [10], with 0.1mm resolution in the Z-axis for the details of the fitting parts, 2 layers for the outside shell, and for infill 10% is recommendable. An impact resistance test is still needed to ensure the reliability of the backpack and to ensure continuous operation of the robot.

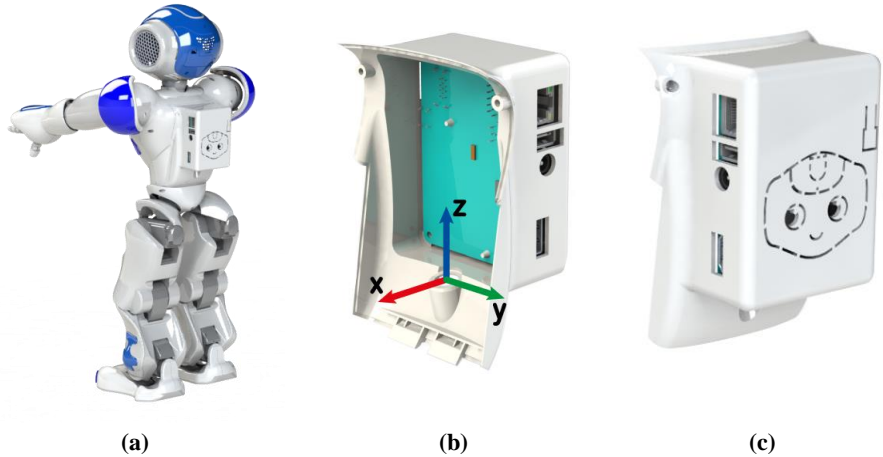

(a)　　　　　　　　(b)　　　　　　　　(c)

**Fig. 1.** A render of the backpack's CAD model. **(a)** Full NAO wearing the backpack. **(b)** Interior view with the ODROID XU4 with coordinate system. **(c)** External view, showing some ports and ventilation holes.

**Stability and mobility.** Given the addition of the backpack to the robot, we expect changes in its stability and mobility; in fact, the center of mass of the torso link moves about 15mm backwards. We tested the basic gait implemented in the B-Human framework on the NAO -previously modifying the inertial configuration of the robot with the new mass and center of mass-, and there were no significant differences with the original gait.

### 2.2 Hardware & Software

The backpack's hardware mainly consists of a development board plus the electronics to power it autonomously.

**ODROID XU4.** In this version of the NAO backpack, we choose a Hardkernel ODROID XU4 as main computation board. Despite existing several other alternatives in the market, we preferred this because its relation performance/price as well as the successful experience shown by the aerial robotics community [11,12]. In addition, XU4 support eMMC modules that surpass considerably a SD-UHS card performance, an ideal feature for robotics applications. A comparison between different boards can be found in Table 1.

**Battery.** Despite the battery available in the NAO kit nominally provides 60 min of active use, in robot soccer we have noticed a performance below this time close to 30 min. ODROID XU4 board requires a 5V / 4A power supply, so we chose a suita-ble battery to power the ODROID during a time comparable to the NAO while playing. We selected a standard 2-Cell 1000mAh LiPo battery plus an UBEC 5V/3A to power the board. Since we are not connecting any other device than a Wi-Fi dongle for specific applications, these specifications are enough for our purposes.

**Connectivity with the robot.** The NAO V5 robot has one USB port as well as Ethernet Gigabit and Wi-Fi 802.11 a/b/g/n connectivity [20]. We are interested in high throughput applications such as image streaming from the robot to the backpack at frame-rate, so we selected the Ethernet connection.

Table 1. Comparison of modern development boards that support Ubuntu Linux and ROS.

| Board | CPU | GPU | RAM | Size | Price |
|---|---|---|---|---|---|
| **Hardkernel ODROID XU4 [13]** | ARM Cortex-A15 32-bit 2GHz x 4 + ARM Cortex-A7 32-bit 1.4GHz x 4 | Mali-T628 MP6 | 2Gb | 83 x 58 mm | $59 |
| **Hardkernel ODROID C2 [14]** | ARM Cortex A53 64-bit 1.5GHz x 4 | Mali 450 | 2Gb | 85 x 56 mm | $46 |
| **Raspberry Pi 3 [15]** | ARM Cortex A-53 64-bit 1.2GHz x 4 | VideoCore IV | 1Gb | 85 x 49 mm | $35 |
| **BeagleBone Black [16]** | Sitara AM335x 1GHz x 1 | SGX530 3D | 512 Mb | 86 x 53mm | $45 |
| **Qualcomm DragonBoard 410c [17]** | ARM Cortex A53 64-bit 1.2 Ghz | Adreno 306 | 1Gb | 85 x 54 mm | $75 |
| **Nvidia Jetson TK1 [18]** | ARM Cortex-A15 32-bit x 4 | Kepler 192 cores | 2Gb | 127 x 127mm | $192 |
| **Nvidia Jetson TX1 (module) [19]** | ARM Cortex-A57 64-bit x 4 | Maxwell 256 cores | 4Gb | 87 x 50 mm | $304 |

### 2.3 Software Architecture

As we stated before, our software architecture is focused in communicating the B-Human framework currently used in the NAO with ROS. A general overview of our approach is presented in Figure 2.

**NAO software: Backpack Communication Modules.** On the NAO side, we use the B-Human framework (BH) [21], which we review briefly. This basically provides two threads: The *Cognition* thread that runs at 30Hz and performs image processing, modeling, localization and decision making; the *Motion* thread reads joints and IMU data, estimates the torso state and computes the robot gait. Each task performed by a thread is named *module*, whilst the information a module provides is called *representation*. We implemented a module in both *Cognition* and *Motion* (**CognitionBackpackComm** and **MotionBackpackComm**) that transmit selected representations from each thread via UDP communication. Data is serialized with BH's libraries, packed and send through different ports to distinguish the source.

In general, all *Motion* data is sent in single packets because it size is below the maximum allowed by the network. However, a critical situation can occur in *Cognition* while sending images because their size is over the maximum; in that case, we send an initial packet with the expected image size, then we fragment and send the image packets sequentially.

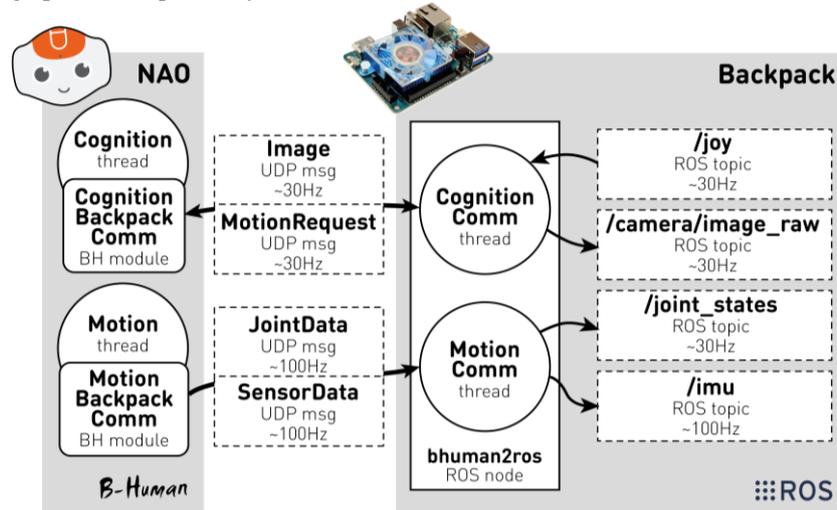

**Fig. 2.** The NAO Backpack's B-Human-to-ROS communication overview with some examples of nodes running on ROS.

**Backpack software: bhuman2ros Bridge.** On the Odroid XU4, we implemented a ROS node **bhuman2ros** that works as a *driver* for the incoming NAO data. The node launches two threads, *CognitionComm* and *MotionComm*, which wait and publish information obtained from *Cognition* and *Motion* respectively, to the corresponding ROS topics. For instance, the BH's *Image* representation is published in ROS as a *sensor_msgs/Image* message in a */camera/image_raw* topic.

Image packets reception faces similar issues than mentioned before while sending them. Since we fragment the image and UDP connection is unreliable, packets can be lost or delayed between the NAO and the ODROID; here we choose a naïve policy of dropping the current image if we receive the initial packet of the next one. In practice, this policy and the direct wire connection between both boards allows us to have a continuous streaming, publishing ROS messages close to sensor rate.

In addition, we implemented a simple joystick mapper in the *CognitionComm* that subscribes the */joy* topic. Joystick commands are translated into BH's *MotionRequest* representations used for gait orders, serialized and send back to NAO using the same approach as used in the Backpack modules.

# 3  Applications

In this section, we present some applications where we tested the NAO with its backpack. The backpack allowed us to run a diversity of software with NAO data, exploiting the facilities that ROS provides. We use an ODROID XU4 with a 16Gb eMMC card for operating system boot and data storage; we did not use a swap area because of disk storage limitations. The operating system was Ubuntu 16.04 MATE with ROS Kinetic.

**Dataset recording.** An essential task in robotics is related to data logging. B-Human's framework provides a custom logging system to record robot data while playing. However, it is very difficult to save a real-time full image streaming because of the limited memory and processor power. While using the backpack, we can record real-time logs of our robots while playing, obtaining images of 320x240 pixels in YUV422 encoding close to 30fps.

**Visual SLAM.** Visual SLAM and Visual Odometry are *hot* research topics in robotics because they allow robots to localize and building maps robustly in an inexpensive way. Nowadays, several open-source systems are available, being ORB-SLAM2 [22], LSD-SLAM [23], SVO [11] and DSO [24] the most popular one. We tested ORB-SLAM2 in the backpack following the official instructions for compiling and execution, being able to run the system at 12fps with images of 640x480 pixels without disabling functionalities, while displaying the GUI via VNC through Ethernet connection.

**Deep Learning Applications.** SegNet [25] and Faster R-CNN [26] are two popular deep networks for semantic segmentation and object detection respectively, so we wanted to test the feasibility of running these algorithms with NAO data. We build the libraries following the official instructions of each package before checking appropriate linking of ARM-based libraries during compilation and disabling GPU features. For SegNet we use caffe-SegNet [27] with pretrained models already included in the repository, particularly we tested the *SegNet SUN low resolution* architecture trained for indoor scenes with 37 different classes; other complex architectures were not tested because of memory issues (this might be possible by adding a swap space). We implemented a ROS node to subscribe and resize the images to fit the 224 x 224 input of SegNet, to measure the inference time and to display the output.

Faster R-CNN was tested in a similar manner. We use the Python implementation from the author [28] with a *Zeigler and Fergus 5-layer (ZF-5)* architecture pretrained with the PASCAL VOC Challenge 2007 dataset; there is an option to try a VGG16 architecture but it was unable to run because of memory as well. We also implemented a test node but since this network resizes the input images automatically we just measure the inference time and displayed output images on demand.

It is important to mention that we also tried some optimizations to improve the inference time. We noticed that changes in the numeric libraries can have a huge impact on performance since they can exploit multi-core parallelism in the ODROID. For instance, changing ATLAS libraries by OpenBLAS can improve performance by almost 50%. Some results are shown in Tables 2 and 3:

**Table 2.** Average processing time results for SegNet SUN low resolution (224x224pixels) with different numeric libraries and scenes.

| Class | ATLAS | OpenBLAS |
|---|---|---|
| **Bathroom** | 26.48s | 13.48s |
| **Rest space** | 24.71s | 13.54s |
| **Living room** | 24.95s | 13.67s |
| **Library** | 25.06s | 13.46s |

**Table 3.** Average processing time results for Faster R-CNN (ZF-5) with different numeric libraries and input image size.

| Size (pixels) | ATLAS | OpenBLAS |
|---|---|---|
| **200x150** | 6.05s | 2.92s |
| **500x375** | 16.14s | 8.03s |
| **800x600** | 20.83s | 11.60s |

## 4 Summary and Future Perspectives

We finish by summarizing the advantages of the NAO Backpack for the SPL, as well as providing guidelines for further developments with this platform, that we hope will help further developments of the league. The 3D printable file of the backpack and the software and hardware specifications to replicate the functionalities of the backpack, will be made available online after this paper is accepted in the symposium.

What we have today with this contribution:

- **Replicability of the platform:** By providing the CAD models as well as the software bridges, we hope that many teams will take advantage of this platform to improve their own research.

- **Testing of popular algorithms in robot soccer domain:** Since the Backpack runs ROS, we can test algorithms already developed within this framework. Hence, both new and old algorithms that are usually prohibitive with the current hardware can be evaluated *without pain* in the robot soccer problem.

- **Collaboration in a *common tongue*:** ROS provides many tools and is supported by a huge community worldwide. By having access to tools such as *rosbag* or *Rviz*, we can record and share real-time recorded datasets of our robots, which can accelerate the development and collaboration in the league.

On the other hand, we foresee the following improvements to our platform:

- **Hardware Upgrades:** The ODROID XU4 is one of the powerful CPU development boards available in the market and widely used in aerial robotics because of its balance between computational power and weight. However, it is not the best choice for computer vision applications since they require high levels of parallelism, such as deep learning. Modern small boards with GPUs, such as the NVIDIA Jetson TX1 would suit better for these applications; it would be interesting to work in new backpacks with GPU support.

- **New bridges for different frameworks:** We covered the problem for the B-Human's framework because of its popularity, and since the ROS bridges already exist for NaoQi [29]. However, to reach compatibility across all the SPL teams, we need ROS bridges for the UPenn's Lua-based framework, UNSW's, Leipzig's, Austin Villa's, and many others.

## Supplementary Material

The NAO backpack package is freely distributed in the UChile Robotics GitHub; this provides the CAD models, list of components, and instructions to run the backpack and NAO software: https://github.com/uchile-robotics/nao-backpack

## Acknowledgments

We thank the B-Human SPL Team for sharing their code-release, providing the communication libraries and data structures used in this project, as well as the ROS developers and community for their efforts to unify robotics developments. This research was partially funded by FONDECYT Project 1161500.